\documentclass[10pt,twocolumn,letterpaper]{article}

\usepackage[pagenumbers]{cvpr} %

\makeatletter
\@namedef{ver@everyshi.sty}{}
\makeatother
\usepackage{tikz}
\usepackage{graphicx}
\usepackage{amsmath}
\usepackage{amssymb}
\usepackage{booktabs}

\usepackage[pagebackref,breaklinks,colorlinks]{hyperref}

\usepackage[capitalize]{cleveref}
\crefname{section}{Sec.}{Secs.}
\Crefname{section}{Section}{Sections}
\Crefname{table}{Table}{Tables}
\crefname{table}{Tab.}{Tabs.}

\usepackage[font=small,labelfont=bf]{caption}
\usepackage{subcaption}
\usepackage{multirow}
\usepackage{graphicx}
\usepackage{epsfig}
\usepackage{tabularx}
\usepackage{mathtools}
\usepackage{tabulary}
\usepackage{booktabs}       %
\usepackage{nicefrac}       %
\usepackage{microtype}      %
\usepackage{booktabs}
\usepackage{pifont}

\usepackage{times}
\usepackage{epsfig}

\usepackage[utf8]{inputenc}
\usepackage{graphicx,verbatimbox}
\usepackage{tcolorbox}

\usepackage{tikz}
\usetikzlibrary{shapes.geometric, arrows, arrows.meta}
\usepackage{pgfplots}
\pgfplotsset{width=7.5cm,compat=1.12}
\usepgfplotslibrary{fillbetween}

\newcommand{\cmark}{\ding{51}\xspace}%
\newcommand{\xmark}{\ding{55}\xspace}%

\begin{document}

\title{Vision and Structured-Language Pretraining for Cross-Modal Food Retrieval} %

\author{
\begin{minipage}{\linewidth}
\begin{center}
 Mustafa Shukor \hspace{0.35cm}  Nicolas Thome  \hspace{0.35cm}
Matthieu Cord \\[0.5cm] 
\scalebox{1.}{Sorbonne University }\\[0.1cm]
\scalebox{1.}{\small{\texttt{\{firstname.lastname\}@sorbonne-universite.fr}}}\\[1cm]
\end{center}
\end{minipage}
}

\maketitle

\begin{abstract}
Vision-Language Pretraining (VLP) and Foundation models have been the go-to recipe for achieving SoTA performance on general benchmarks. However, leveraging these powerful techniques for more complex vision-language tasks, such as cooking applications, with more structured input data, is still little investigated.
In this work, we propose to leverage these techniques for structured-text based computational cuisine tasks.  Our strategy, dubbed VLPCook, first transforms existing image-text pairs to image and structured-text pairs. This allows to pretrain our VLPCook model using VLP objectives adapted to the strutured data of the resulting datasets, then finetuning it on downstream computational cooking tasks. During finetuning, we also enrich the visual encoder, leveraging pretrained foundation models (\textit{e.g.} CLIP) to provide local and global textual context. VLPCook outperforms current SoTA by a significant margin (+3.3 Recall@1 absolute improvement) on the task of Cross-Modal Food Retrieval on the large Recipe1M dataset. We conduct further experiments on VLP to validate their importance, especially on the Recipe1M+ dataset. Finally, we validate the generalization of the approach to other tasks (\emph{i.e}, Food Recognition) and domains with structured text such as the Medical domain on the ROCO dataset. The code is available here: \href{https://github.com/mshukor/VLPCook}{https://github.com/mshukor/VLPCook}.
\end{abstract}

\section{Introduction}
\label{sec:intro}

Vision-Language Pretraining (VLP) \cite{chen2020uniter, tan-bansal-2019-lxmert, su2019vlbert, li2021albef, dou2021empiricalmeter} has become the general recipe to attain SoTA results on downstream unimodal and multimodal tasks, with the key success is learning a shared latent space where all modalities are aligned. This paradigm generally helps to overcome the human labor associated with designing a task or domain customized approaches, and pushes towards more simplification, by unifying the model, training objective and input/output format \cite{wang2022imagebeit3, wang2022unifyingofa, chen2022pali}. As going large scale is an important ingredient to push the performance limits, we have witnessed recently a lot of work going in this direction, leading to what so-called foundation models \cite{alayrac2022flamingo, yu2022coca, radford2021learning, jia2021scaling_align, li2022blip, chen2022pali}.

However, these approaches are still evaluated on simple downstream tasks, to the detriment of more complex albeit important tasks. The current evaluation schema considers tasks such as VQA \cite{antol2015vqa}, Visual entailment \cite{xie2019visualsnli}, Image-Text Retrieval \cite{plummer2015flickr30k}, Image Classification and other general benchmarks that highly resemble the pretraining data, in terms of image distribution, text format, length and structure.
Similarly, existing Foundation models have shown great transfer capabilities to several downstream tasks, however, it is still also unclear how they perform beyond common tasks. The key stumbling block to leverage VLP and Foundation models for such domains, is the complex input that is hard to digest. In particular the tasks involving images with associated text that goes beyond simple image caption, to richer, longer and structured text.

In this work, we question how to leverage VLP and existing Foundation models for tasks requiring structured text. As image-text alignment has proven to be successful for multimodal tasks, we focus on Image-Text Retrieval being one of the best benchmarks to evaluate such alignment. To validate the proposed approach, we consider the traditional task of on Cross-Modal Food Retrieval \cite{Salvador_2017_CVPR_recipe1m},  aiming at bridging the gap between VLP and Computational Cooking.

\begin{figure*}[t]
    \centering
    \includegraphics[width=\linewidth]{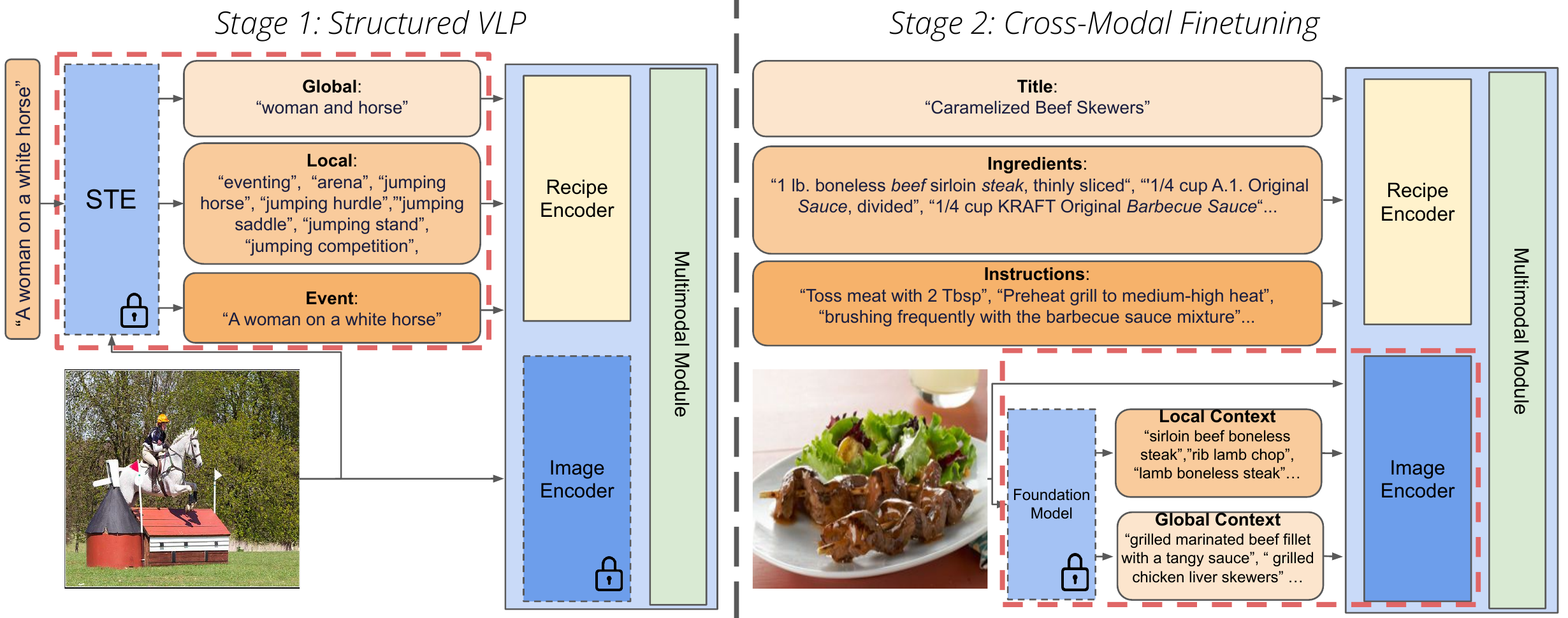}
     \caption{
     \textbf{VLPCook framework with 2 sequential stages}. Stage 1 (left) or VSLP (Sec.~\ref{sec:structured_vlp}): the Structured Text Extraction (STE) module transforms the caption to a structured recipe-like input that is used to pretrain the model on a large corpus of structured text and images. Stage 2 (right) or Cross-Modal Finetuning (Sec.~\ref{sec:finetuning}): we leverage existing foundation models to enrich the vision encoder with local and global textual context. Main contributions are highlighted in red. The lock symbol means the model is frozen.
     }
    \label{fig:main}
\end{figure*}

Computational Cooking or Food applications \cite{categ, percept, recommendation, Salvador_2017_CVPR_recipe1m} are one of the important applications that fit very well in this marginalized list, with no existing work to bridge the gap with VLP. In particular, Cross-Modal Food Retrieval \cite{Salvador_2017_CVPR_recipe1m, carvalho2018adamine, salvador2021revamping, shukor2022transformertfood} which has gained a lot of attention in the recent years and is the current main benchmark to assess the model performance on computational cooking. The images are of different food plates with high inter and low intra category similarity. The text, consists of the corresponding recipe that is composed of 3 entities; title (global description), ingredients (local descriptions, objects or entities that might be seen or not) and instructions (events that we generally see only their effects or final results).

As the main hurdle to enable VLP for food models is the input data, we choose to adapt the input data to be compatible, structurally and semantically, to some extent, to fit in these models. In addition, and pushing on the environmentally responsible idea of reusing existing models, we exploit existing large scale Vision-Language Models (VLMs), to guide the vision encoder with structured context. This guidance is through region-level or local context (\emph{e.g.} ingredients), and image-level or global context (\emph{e.g.} titles). Our approach, dubbed VLPCook, consists of 2 stages; (1) Vision and Structured-Language Pretraining (VSLP) of the model on the created structured text, then (2) Cross-Modal Finetuning guided by foundation models. The approach is illustrated in Fig.~\ref{fig:main}.

Our main contributions can be summarized as follows:
\begin{itemize}
    \item We propose a new approach for transforming existing datasets of image-text pairs to datasets of image and structured-text pairs, and show that VLP on such datasets gives significant improvement.  
    \item We propose a new model that leverages existing pretrained foundation models to inject structured local and global textual context to guide the visual encoder. 
\end{itemize}

To validate the work, we conduct an extensive experimental study on the challenging task of Cross-Modal Food Retrieval, which leads to the following interesting outcomes:
\begin{itemize}
    \item VLPCook outperforms significantly other SoTA on the Recipe1M dataset, with absolute improvement of +3 and +3.3 of R@1 on the 1k and 10k setups respectively.
    \item The first work showing the effectiveness of VLP in the cooking context, after experimenting with different kinds of existing food approaches.
    \item Despite what was reported \cite{marin2019recipe1m+} on the poor generalization from Recipe1M+ to Recipe1M, we show that pretraining on this large dataset can unlock its potential, and lead to large improvement of +2.4 R@1 on Recipe1M test set.
    \item Contrary to recent findings showing that foundation models can attain SoTA on standard benchmarks (\emph{e.g.} VQA v2, COCO retrieval), we show that finetuning these models lag significantly behind SoTA on the underlying task of Cross-Modal Food Retrieval.
    \item We validate the generalization of the work to other tasks (\emph{i.e.}, Food Recognition) and domains, such as the Medical domain, showing significant improvement over baselines.
    
\end{itemize}

\section{Related Work}
\label{sec:related}
\noindent\textbf{Vision and Language Pretraining (VLP)}
Vision and Language Pretraining (VLP) \cite{chen2020uniter, tan-bansal-2019-lxmert, su2019vlbert} aims at learning vision-language representation by pretraining on datasets of images and texts (\cite{Sharma2018ConceptualCA, sbu, schuhmann2021laion, radford2021learning, alayrac2022flamingo}). The model is then evaluated on several downstream tasks such as VQA \cite{antol2015vqa}, NLVR2 \cite{suhr-etal-2019-corpusnlvr}, image-text retrieval \cite{plummer2015flickr30k}  and image captioning \cite{lin2014microsoftcoco}. 
This line of research has shown promising success in the last few years, leading to state of art (SoTA) results \cite{li2021albef, dou2021empiricalmeter, li2022blip} compared to task-customised models, and providing modular encoders that are seamlessly used in a variety of ways. Besides several other improvements, the major ones have been either in the architectural design, or the pretraining objectives. On the model side, we have models with separate vision and language encoders, without significant cross modal interaction (\emph{e.g.}, CLIP \cite{radford2021learning}, ALIGN \cite{jia2021scaling}). Despite their fast inference, they are data hungry and perform poorly on tasks that need deeper reasoning. To overcome these limitations, heavy fusion models use a cross modal interaction module \cite{kim2021vilt, lu2019vilbert, li2019visualbert, chen2020uniter, su2019vlbert, li2020unicoder, zhang2021vinvl} which is added on top of unimodal encoders \cite{meter, yang2022visiontriplecont, li2021albef, vicha} leading to hybrid models. These hybrid approaches have succeeded to get SoTA results while training on reasonably sized datasets.  On the learning side, the main training objectives can be categorised into contrastive (ITC \cite{radford2021learning}, ITM \cite{chen2020uniter}) and masked predictions (MLM \cite{devlin2018bert}, MIM \cite{vicha, dou2021empiricalmeter}). The models that work best are those that combine several objectives, however, at large scale, there are many attempts to unify pretraining tasks.  

\noindent\textbf{Leveraging Foundation Models}
Foundation models \cite{radford2021learning, singh2021flava, yu2022coca, alayrac2022flamingo, wang2022unifyingofa, wang2022imagebeit3} draw some similarity with VLP, however here the objective is to develop a general model that can  be adapted to many unimodal and multimodal tasks. Here there is more emphasis on large scale, in terms of tranining data \cite{radford2021learning}, and model size \cite{yu2022coca} and on unification of the architectural design and training objectives \cite{wang2022unifyingofa, wang2022imagebeit3}. In spite of being successful, due to the need for huge resources to traininig these models from scratch, researchers and practioners have leveraged them, without the burden of retraining; such as initialization and finetuning \cite{shen2022how, shukor2022transformertfood}, as frozen modules \cite{ramesh2022hierarchicaldalle2, couairon2022flexit, su2022languagemagic}, enriching the input \cite{sara2022retrieval} and extracting visual concepts \cite{vicha}. In our work, we leverage existing pretrained foundation models to extract different aspects of textual contexts to enrich the visual representation.

\noindent\textbf{Food Applications and Learning from Sructured Data}
Many work have been proposed in the recent years for food tasks, such as food categorization \cite{food101}, calorie estimation \cite{foodcalorie}, image generation \cite{cookgan} and cross modal retrieval \cite{Salvador_2017_CVPR_recipe1m}. Since the inception of large scale food datasets such Recipe1M \cite{Salvador_2017_CVPR_recipe1m} followed by Recipe1M+ \cite{marin2019recipe1m+} the task of cross-modal retrieval have gained a lot of attention. In terms of performance and architectural designs, cross modal food retrieval work can be divided into transformer-based \cite{salvador2021revamping, guerrero2021cross_xmrs, shukor2022transformertfood, programfood} or transformer-free \cite{Salvador_2017_CVPR_recipe1m, carvalho2018adamine, dac, wang2019learning_acme, wang2021cross_scan, Zhu_2019_CVPR_r2gan} approaches, with a significant improvements of the former. Specifically, on the vision side, ViT \cite{dosovitskiy2021anvit} is used as an image encoder, and on the recipe side, standard \cite{guerrero2021cross_xmrs} or hierarchical transformers \cite{salvador2021revamping, shukor2022transformertfood} are adopted. In terms of training objectives, almost all approaches use triplet loss \cite{NIPS2005_a7f592ce_triplet, ding2015deep, schroff2015facenet} in addition to some regularization such as semantic triplet \cite{carvalho2018adamine, shukor2022transformertfood}, embedding classification \cite{Salvador_2017_CVPR_recipe1m}, adversarial losses \cite{wang2019learning_acme} and multimodal regularization with image-text matching objective \cite{shukor2022transformertfood}.  
In addition to food applications, learning from structured texts and images has been investigated in several domains and tasks, such as  Medical applications \cite{pelka2018radiologyroco}, News applications \cite{biten2019goodnews}, Multimedia Event extraction \cite{li-etal-2020-crossmee, li-etal-2020-gaia} and Situation Recognition \cite{Suhail_2019_ICCV_situation, Cooray_2020_CVPR_situation}. 
In the context of VLP, few work have been recently proposed \cite{li2022clipevent, li2022mvp}, however, they do not consider the case of structured text as input during test and focus on learning a structural representations.

\section{VLPCook}
\label{sec:method}

\noindent\textbf{Overview:} We introduce VLPCook, the first work trying to bridge the gap between VLP and the Computational Cooking domain. VLPCook proposes a novel pretraining pipeline that solves the issues of complex cooking inputs, and a finetuning framework that leverages this pretraining and foundation models for cooking tasks, such as the task of Cross-Modal Food Retrieval. VLPCook consists in 2 stages: (1) Vision and Structured-Language Pretraining (VSLP in Sec.~\ref{sec:structured_vlp}); to perform VLP relevant to complex cooking recipes, we transform the image captions (in existing image-text pairs datasets) to structured text, and form new datasets of image and structured text pairs. This allows us to benefit from a large-scale VLP adapted to the specificity of cooking datasets. (2) Cross-Modal Finetuning (Sec.~\ref{sec:finetuning}); on the downstream cooking task, where we leverage existing foundation models, without any retraining, to contextualize the visual encoder with local and global textual context. The approach is illustrated in Fig.~\ref{fig:main}. 
As our goal is to leverage VLP and foundation models and show their benefits for the cooking domain, we decide to build our approach on top of recent SoTA food models and keep as much as possible the same model architecture/finetuning objectives.

\noindent\textbf{Background on VLP:}
VLP consists of pretraining Vision-Language models on large datasets of image-text pairs, then finetuning on several multimodal downstream tasks. Several pretraining objectives are used in VLP. Here we focus only on 2 of them; Image-Text Contrastive (ITC) and Image-Text Matching (ITM):

\noindent\textit{ITC:} \label{sec:itc} several ITC losses have been proposed, such as InfoNCE \cite{oord2018representation, zhang2020contrastive, deepmetric} and triplet loss \cite{ding2015deep, NIPS2005_a7f592ce_triplet}. In this work, we use a triplet loss on top of the unimodal encoders. On one hand, we pull the image embedding to be close to the corresponding recipe embedding, and vice versa, and on the other hand, we push far away the embeddings of different recipes. ITC is used to globally align both modalities, which is important for tasks such as cross-modal retrieval.

\noindent\textit{ITM:} \label{sec:itm} is a binary classification loss to train the model to predict matched image-text pairs \cite{chen2020uniter}. This loss is applied on top of the multimodal module (\emph{e.g.}, transformer decoder) and aims to learn more fine-grained interaction between modalities. 

\begin{figure}[t]
    \centering
    \includegraphics[width=\linewidth]{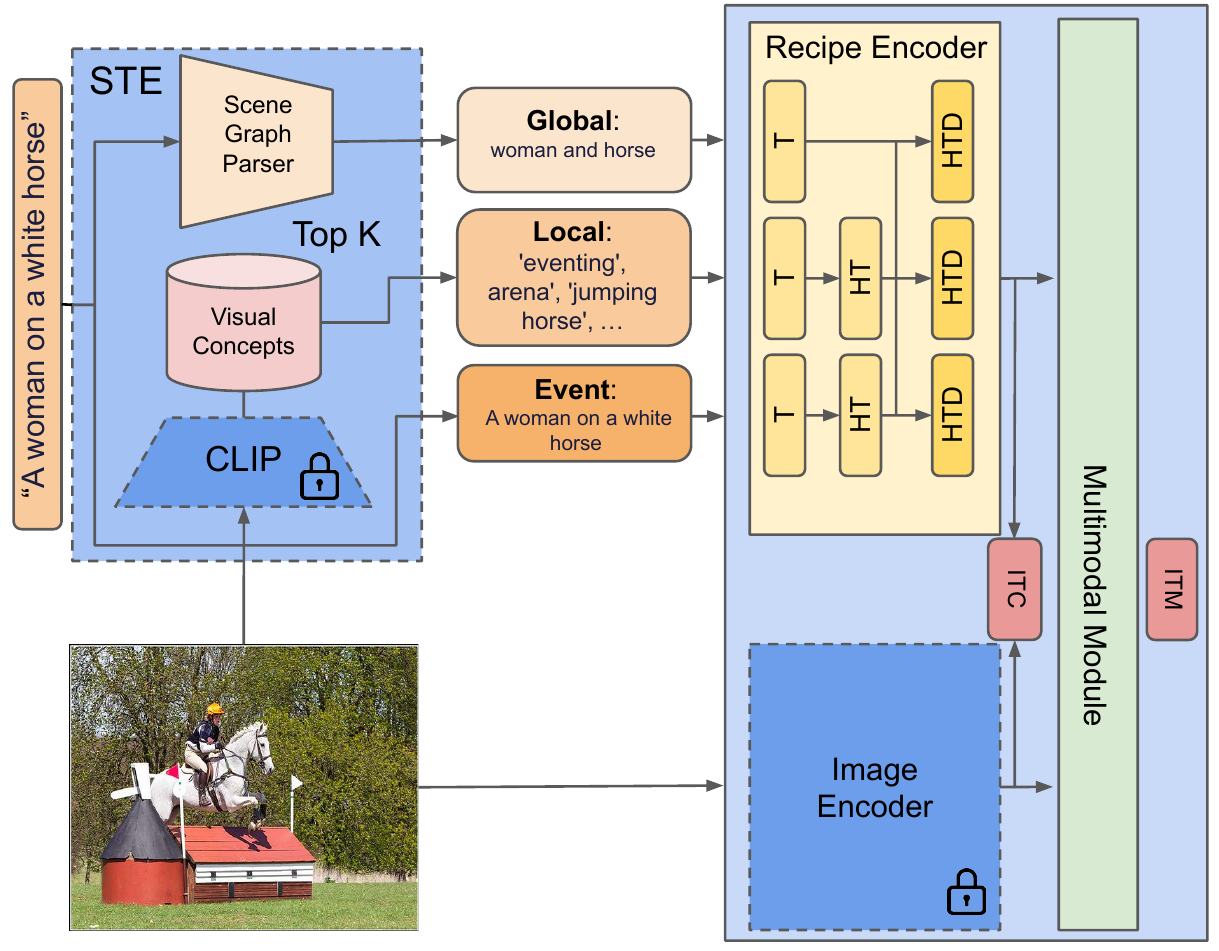}
     \caption{\textbf{Illustration of our VSLP (Stage 1 of VLPCook)}. To enable VLP for food models, image-text pairs are transformed to image and structured-text pairs, that are compatible with hierarchical recipe encoders. The Structured Text Extraction (STE) module generates 3 entities; (a) global description ("title") using SGP, local descriptions ("ingredients") using CLIP-based retrieval, and the "event" ("instructions") which can be simply the caption. During VLP, we optimize ITC and ITM losses and keep the vision encoder frozen.}
    \label{fig:vlp_recipe_encoder}
\end{figure}
\subsection{Vision and Structured-Language Pretraining (VSLP)}
\label{sec:structured_vlp}
Existing VLP approaches use image captions; usually a one sentence describing a general event, or the scene in the image. Despite being easily scraped from the internet, and successful in many general downstream tasks, image captions are not directly aligned with some domains such as Food applications. Specifically, image-captions generally contain one sentence describing globally the image, while recipes are longer ($> 200$ words), with a richer description, including global (title), local (ingredients), and structured (hierarchical) information. 

Here we focus on computational cooking tasks that require such complex text input. The text or the recipe consists of different elements, forming a hierarchical structure; global information about the image (\emph{e.g.}, title), local information (\emph{e.g.} ingredients) and the interaction between different entities (\emph{e.g.} instructions). The text is long (\emph{e.g.} more than 10 ingredients/instructions) and rich, as it contains very specific details (\emph{e.g.} ingredients name and quantity). Recent food models have dedicated recipe encoders \cite{salvador2021revamping, shukor2022transformertfood} to exploit such structure. They use several stages of transformers: one for each ingredient/instruction (T), another for the list of ingredients/instructions (HT), and the last stage with transformer decoders (HTD) that take the tokens of one entity as query and the tokens of other ones as keys and values (Fig.\ref{fig:vlp_recipe_encoder}).

To bridge this gap between VLP and the food domain, we propose first to create datasets of structured image-text pairs, then use them to pretrain food models. This stage is illustrated in Fig.~~\ref{fig:vlp_recipe_encoder}.

\paragraph{From Image Captions to Structured Text (Recipe-fying the captions):}
\label{sec:transform_captions}
we propose a new approach to transform existing image captions, in existing datasets of image and text pairs, to richer and structured text. Transforming existing datasets helps us to leverage large scale ones, which is cheaper than creating large scale datasets of image-recipe pairs from scratch.
We make the analogy between the obtained text and recipes and detail the process in the following:

\noindent\textit{Global information (Title):} we assume that the caption describes either the global scene or the main event in the image, and use it to extract the title. However, it may also include some unnecessary details to be considered for the title, as well as noise (especially for datasets scraped from internet). As a way to filter out the caption and keep the main elements, we extract only the objects using Scene Graph Parsing (SGP) \cite{schuster-etal-2015-generating_sgp} techniques and assemble them with a simple "and" (\emph{e.g.}, title: Woman and Piano and stage).

\noindent\textit{Local information (Ingredients):} here, local entities or objects in the image should be included. Relying on the caption alone is not optimal, as it contains only few seen objects, besides referring to global aspects of the scene. On the other hand, we do not want to be limited to seen objects and include unseen but relevant objects, which is the case for ingredients in food tasks (\emph{e.g.} salt, sugar). This motivates us to leverage additional sources of information to extract all relevant, seen or unseen, objects. 
To this end, we use existing foundation models, without retraining them, as they enjoy good generalization capabilities on different domains and tasks, to retrieve the closest entities. 
Specifically, these entities are retrieved from a database that contains all objects extracted from the captions of several image-text datasets (\emph{e.g.} COCO, SBU). To get the local entities of an image, the image is fed to a CLIP visual encoder \cite{radford2021learning}, then a cosine similarity is applied to compute the distance between the image  and all textual embeddings of local entities, to select the closest k ones.

\noindent\textit{Event (instructions):}
To describe the event, we consider the caption. Even though the caption might describe only one event in which some of the objects participate, we found that using additional captions does not help significantly.

Note that, this approach can be leveraged in a straightforward way to other domains with structured text, such as Medical applications. 
\paragraph{VLP with Structured Text:}
Once we create datasets of images and structured-text pairs, we can feed such data to the hierarchical text encoder and pretrain our model (Fig.~\ref{fig:vlp_recipe_encoder}) using standard VLP objectives. We use both ITC and ITM objectives. For text-to-image ITC loss (similarly for the image-to-text ITC), the triplet loss is fed with the text ($t$) and image ($v$) embeddings:  
 \begin{align}
 \label{eq:triplet_structured}
    & l(t_a, v_p, v_n, \alpha)  = [d(t_a, v_p) +\alpha - d(t_a, v_n)]_+, \\ \nonumber
     & t = \mathcal{E}_t(G,L,E), \quad v = \mathcal{E}_v(I),
\end{align}

where $t_a$, $v_p$ and $v_n$ are the anchor, positive and negative embeddings respectively, $\alpha$ is the margin and $d(\cdot, \cdot)$ is a distance function. 
The image embedding is obtained after processing the image (I) with the image encoder $\mathcal{E}_v$. The text embedding is obtained after processing the structured text, with the extracted local ($L$), global ($G$) and  event ($E$) elements. Specifically, $\mathcal{E}_t$ first encodes each entity independently using transformer encoders, then exploits their interactions with cross attention \cite{shukor2022transformertfood}. We then compute ITC loss ($\mathcal{L}_{itc}$) by summing the triplet losses over the batch and weight the loss by the inverse of number of active triplet as done in Adamine \cite{carvalho2018adamine}. All examples in the batch are considered negatives, except the images that correspond to the recipe and vice-versa. The ITM loss can be written as:
\begin{align}
\label{eq:itm}
    \mathcal{L}_{itm} & = -\mathbb{E}_{T, I \sim D}[y\log(s(T, I)) + \\ \nonumber 
    & (1-y)\log(1-s(T, I))],
\end{align}
where $y$ is the label (\emph{i.e.}, 1 for matching pairs and 0 otherwise) and $D$ is the set of structured text ($T=\{L, G, E\}$) and image (I) pairs, and $s()$ is the score on top of the multimodal module. The total loss becomes:\vspace{-0.1cm}
\begin{equation}
    \mathcal{L} = \mathcal{L}_{itc} + \lambda \mathcal{L}_{itm}
\end{equation}

On the image side, to ease the pretraining, and leverage the initial visual representation, we follow LiT \cite{zhai2022lit} and keep the vision encoder frozen, we also find that this gives better results. We use a general vocabulary (used in BERT) and change the embedding layer during this stage.

\begin{figure}[h]
    \centering
    \includegraphics[width=\linewidth]{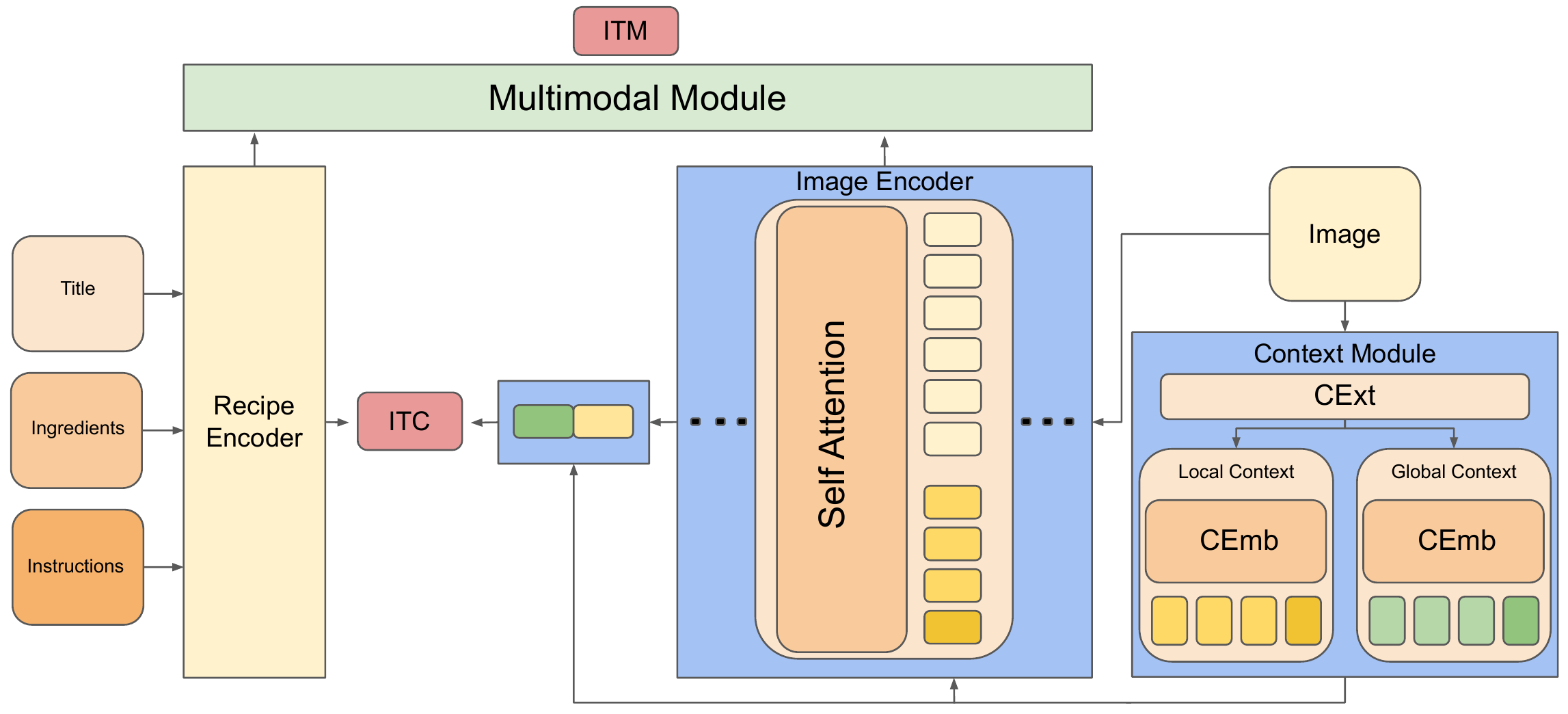}
     \caption{\textbf{Illustration of our contextualized vision encoder (stage 2 of VLPCook)}. The ViT is contextualized by the context module, which extracts local and global context (CExt), then project them using a light-weight module (CEmb) to obtain the context tokens. Local context tokens are concatenated to the image tokens at the input of the ViT, and the global context token (CLS token) is concatenated at the output.}
    \label{fig:VLPCook}
\end{figure}

\subsection{Leveraging Foundation Models for Structured Downstream Tasks}
\label{sec:finetuning}
We propose to leverage foundation models (CLIP \cite{radford2021learning}), without any retraining, for cross modal food retrieval. The approach is based on injecting local and global textual contexts in the image encoder, to enrich the visual representation and steer it towards the textual embedding space. This context inherits the features and biases in the pretrained CLIP, which excels in general cross-modal retrieval tasks.  We adopt a vision transformer (ViT  \cite{dosovitskiy2021anvit}) on the image side. We elaborate first on how we contextualize the ViT, then we detail the finetuning step. The model is illustrated in Fig.~\ref{fig:VLPCook}.

\paragraph{Contextualized Visual Representation:}
\label{sec:vlms_context}
We inject different types of contexts during the image encoding; global and local. For global context, we inject different titles, while for local one, we inject different ingredients. The titles and ingredients are extracted from the image using our CLIP-based retrieval approach (Sec.~\ref{sec:transform_captions}). During training, we inject different titles, ingredients and different combination of them for each batch to add more variability and some regularization during training.

To obtain the context tokens, we concatenate all context elements (all titles for global context or all ingredients for local one) to form one sentence that is embedded using the Context Embedding (CEmb) module (Fig.~\ref{fig:VLPCook}). CEmb consists of a light-weight text encoder and a linear projection layer to project the textual tokens to the space of the visual tokens. We inject the local context early, in the input of the ViT (concatenation to the image tokens), and the global one, later in its output (concatenation of CLS token before the linear projection), where we have higher abstraction level and more global representation. The forward pass of the contextualized ViT can be expressed as follows:
\begin{align}
\label{eq:vit_context}
    x & = ViT(Concat(i_1, .., i_k, c^l_1, .., c^l_p)) \\ \nonumber 
    x &= F(Concat(x_{cls}, c^g_{cls}))
\end{align}
Where $i_j$, $c^l_j$ and $c^g_j$ are the tokens of the image ($k$ tokens), local context ($p$ tokens) and global context respectively. The ${cls}$ means the class token and $F$ is a linear layer.

This is different from other food approaches that add only global information (food category or class) later by concatenating it to the visual embedding \cite{xie2021learning_jema} or other approaches that concatenate object tags (OSCAR \cite{li2020oscar}) or visual concepts (ViCHA \cite{vicha}) only at the input, without any distinction between local and global contexts. Our approach is also inspired by prompt tuning techniques \cite{lester2021powerprompttuning, jia2022vpt, lu2022prompt} where a couple of learnable tokens are concatenated before the main text to adapt the frozen model to a given task.

\paragraph{Finetuning:}
We finetune the model on cross-modal food retrieval. During this stage, we inject the local and global contexts (Sec ~\ref{sec:vlms_context}). The model consists of a ViT, hierarchical recipe encoder and a mulitmodal module \cite{shukor2022transformertfood}, mainly we train the model using Adamine triplet loss \cite{carvalho2018adamine} with incremental margin, in addition to the ITM loss as a multimodal regularization at the output of the mulimodal module. During test, we only use the unimodal encoders for fast retrieval. The context is injected also during test.

\section{Experiments}
\label{sec:results}
In this section we detail the experimental results.

\noindent\textbf{Datasets:}
\label{sec:data}
We use several datasets; such as Recipe1M \cite{Salvador_2017_CVPR_recipe1m} (239 k, 51 k, 51 k pairs as training, validation and test set) where each example consists of a recipe (title, ingredients, instructions) and image pair. Recipe1M+ \cite{marin2019recipe1m+} that is an extension of Recipe1M with 13M images and 1M recipe, and Image and Structured Text pairs (IST), which is our dataset constructed with the STE module from 3 public datasets; COCO \cite{lin2014microsoftcoco}, Visual Genome \cite{krishna2017visualgenome} and SBU \cite{sbu} to form a total of 2M pairs including around 1M different images.

\noindent\textbf{Implementation details:} the model consists of hierarchical transformer encoders and decoders on the recipe side, a ViT-B/16 on the image side and a multimodal module. For VLP, we start by pretraining (with frozen ViT) with learning rate (lr) of 1e-5 and total batch size of 200 on 4 GPUs (50 per GPU) for 30 epochs.  In the second finetuning stage on Recipe1M, we follow the implementation details of other work \cite{shukor2022transformertfood}. We associate each image to 5 titles and 15 ingredients. During training, we sample only 2 titles and 4 ingredients randomly in each batch. The context is embedded by the first 2 layers of the BERT \cite{devlin2018bert} encoder, followed by linear projection (more details in the appendix).

\setlength{\tabcolsep}{7pt}
\begin{table}[h]
\centering
\resizebox{0.48\textwidth}{!}{
\begin{tabular}{@{}l|cccccc@{}}
\toprule
\multirow{3}{*}{}    & \multicolumn{6}{c}{\textbf{10k}} \\ \cmidrule(l){2-7} 
                   & \multicolumn{3}{c}{\textbf{image-to-recipe}} & \multicolumn{3}{c}{\textbf{recipe-to-image}} \\ \cmidrule(l){2-4} \cmidrule(l){5-7} 
                  &  R@1      & R@5    & R@10    &  R@1     & R@5     & R@10    \\ \midrule
Adamine \cite{carvalho2018adamine}       &   14.8     &    34.6    &    46.1 &  14.9    &   35.3    &  45.2   \\
R2GAN \cite{Zhu_2019_CVPR_r2gan}                          &   13.5      &   33.5    &     44.9   &       14.2     &   35.0     &   46.8     \\
MCEN \cite{fu2020mcen}                       &      20.3   &    43.3   &   54.4     &       21.4     &  44.3      &    55.2    \\
ACME \cite{wang2019learning_acme}                 &   22.9     &  46.8     &    57.9    &       24.4      &    47.9    &      59.0  \\
SN \cite{sentencebased}                   &    22.1    &    45.9   &    56.9    &        23.4   &  47.3      &  57.9      \\
IMHF \cite{intra_inter}                    &    23.4    &    48.2   &    58.4    &         24.9   &  48.3      &  59..4      \\
\textit{Wang et. al}  \cite{wang2021learning}                   &    23.4    &    48.8   &    60.1    &        24.6   &  50.0      &  61.0      \\
SCAN \cite{wang2021cross_scan}                 &    23.7    &    49.3   &    60.6    &        25.3   &  50.6      &  61.6      \\
HF-ICMA \cite{hybrid_fusion}                 &    24.0    &    51.6   &    65.4    &        25.6   &  54.8      &  67.3      \\
MSJE \cite{xie2021learning}               &      25.6    &    52.1   &    63.8    &       26.2   &  52.5      &  64.1      \\

SEJE \cite{xie2021learning2}              &    26.9    &    54.0   &    65.6    &        27.2   &  54.4      &  66.1      \\

M-SIA \cite{multisubspace} &     29.2     &    55.0   &    66.2        &     30.3   &     55.6   &  66.5     \\

DaC \cite{dac}             &     30.0     &    56.5   &    67.0        &     -   &     -   &        \\

X-MRS \cite{guerrero2021cross_xmrs}             &       32.9     &    60.6   &    71.2    &      33.0   &     60.4   &  70.7       \\

H-T (ViT) \cite{salvador2021revamping}            &       33.5     &    62.1   &    72.8    &      33.7   &     62.2   &  72.7       \\
T-Food (ViT) \cite{shukor2022transformertfood}          &  40.0          & 67.0  &75.9 &  41.0       &67.3         &      75.9            \\ 
T-Food (CLIP-ViT)  \cite{shukor2022transformertfood}              & 43.4 & 70.7   & 79.7  &    44.6    & 71.2        & 79.7                 \\ \midrule
VLPCook           &      \underline{45.3}  & \underline{72.4}   & \underline{80.8}  &     \underline{46.4}    & \underline{73.1}        & \underline{80.9}                 \\ 
VLPCook  (R1M+)          &     \textbf{46.7}  & \textbf{73.3}   & \textbf{83.3}  &     \textbf{47.8}    & \textbf{74.1}   & \textbf{81.8}                 \\ 
\bottomrule
\end{tabular}
}
\caption{\textbf{Comparison with other work.} Recall@k ($\uparrow$) is reported on the Recipe1M test set. Our approaches (VLPCook) significantly outperform all existing work. Best metrics are in bold, and next best metrics are underlined.}
\label{tab:main}
\end{table}

\subsection{VLPCook Results}
\label{sec:VLPCook_exp}

\paragraph{Results on Recipe1M:}
\noindent\textit{Comparison with SoTA:} Tab.~\ref{tab:main} shows a comparison with existing approaches on the test set of Recipe1M. VLPCook significantly outperforms current SoTA (+1.9 R@1) on the challenging 10k setup. Importantly, the gap between VLPCook pretrained on Recipe1M+ and SoTA is even bigger (+3.4 R@1 on 10k).

\begin{figure*}[h]
    \centering
    \includegraphics[width=0.95\linewidth]{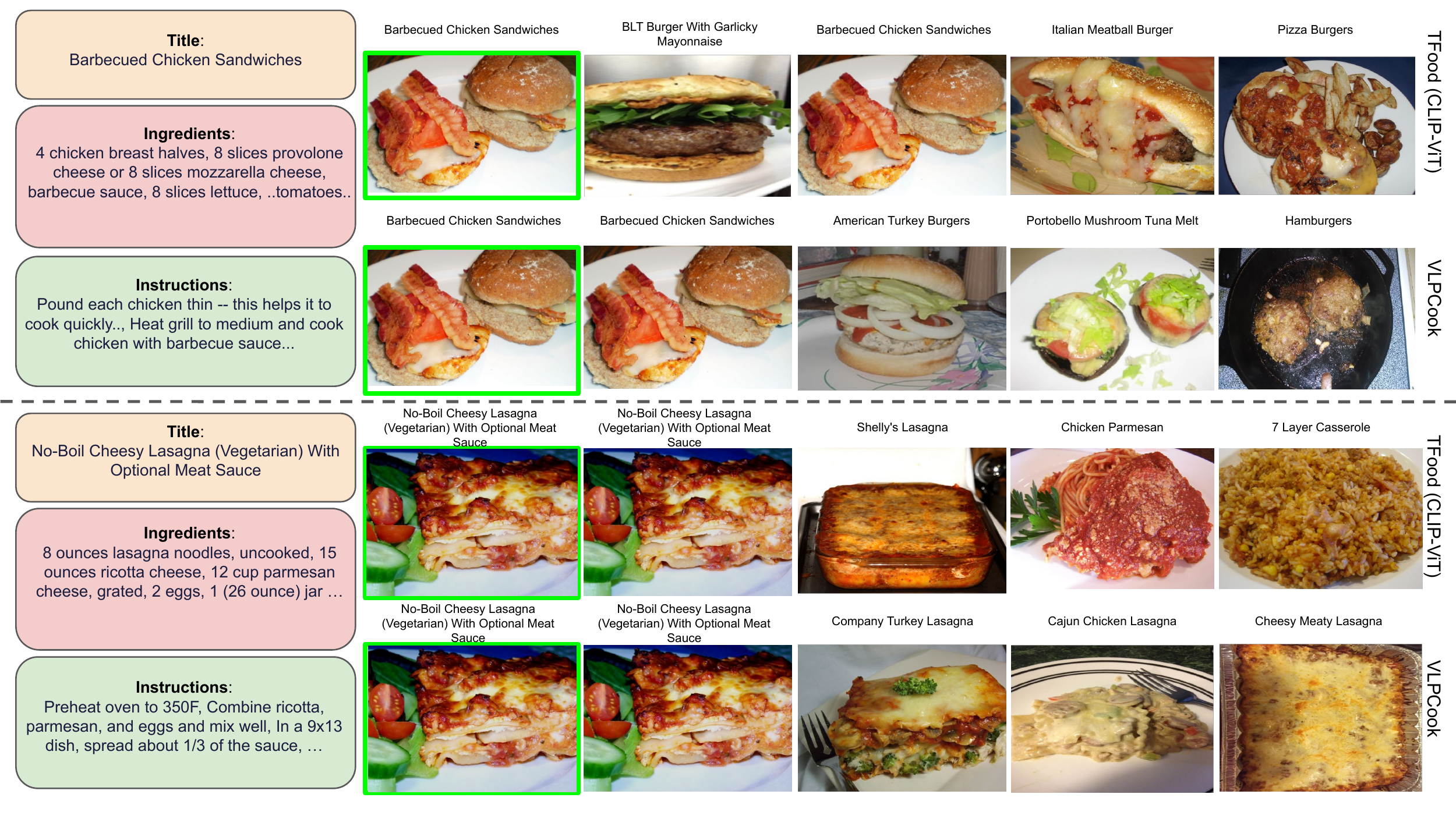}
     \caption{\textbf{Recipe-to-image comparison on the Recipe1M test set, 1k setup.} TFood (first and third rows) {\textit{vs.}} our VLPCook (second and fourth rows). The image in green is the ground truth, followed by the top 4 retrieved images in order. One can notice that our VLPCook approach  better captures some finegrained details (type of meat) and most of the retrieved images are semantically similar.}
    \label{fig:vis_r2im}
\end{figure*}

\noindent\textit{Qualitative Comparison with SoTA:} we show some qualitative results in Fig.~\ref{fig:vis_r2im}. We can notice the superiority of VLPCook compared to the current SoTA (Tfood CLIP-ViT). Specifically, in the first example, VLPCook correctly retrieves the right image. In the second example, our approach retrieves semantically similar images (Lasagna), while for TFood, there are totally different plates (\emph{e.g.} rice, pasta).

\setlength{\tabcolsep}{7pt}
\begin{table}[h]
\centering
\resizebox{0.48\textwidth}{!}{
\begin{tabular}{@{}l|cccccc@{}}
\toprule
                   & \multicolumn{3}{c}{\textbf{image-to-recipe}} & \multicolumn{3}{c}{\textbf{recipe-to-image}} \\ 
                  &  R@1      & R@5    & R@10    &  R@1     & R@5     & R@10    \\ \midrule
Marin et al. \cite{marin2019recipe1m+}        & 17.0     &  38.0     &   48.0     &      17.0    &    42.0  &     54.0      \\
VLPCook$^*$  & \textbf{45.2}     &  \textbf{75.9}     &   \textbf{84.0}     &      \textbf{47.3}    &    \textbf{77.6}  &     \textbf{85.3} \\
                 
\bottomrule
\end{tabular}
}
\caption{\textbf{Comparison with other work.} Recall@k ($\uparrow$) is reported on the Recipe1M+ test set (1k setup). Best metrics are in bold. VLPCook$^*$ here is without VLP.}
\label{tab:main_recipe1m+}
\end{table}
\paragraph{Results on Recipe1M+:} in Tab.~\ref{tab:main_recipe1m+}, we show the first finetuning results on Recipe1M+ with interesting scores (more details in the appendix). Due to the large dataset size, we report the results of VLPCook without VLP (only with the context module). The scores are almost multiplied by 3 compared to the baseline \cite{marin2019recipe1m+}. However, there is a big gap between the scores on this dataset and those on Recipe1M, which makes it more challenging and more interesting to devise more complex approaches in the future.

\begin{table}[th]
\small
\centering
\setlength{\tabcolsep}{5pt}
\resizebox{0.45\textwidth}{!}{
\begin{tabular}{@{}l|ccc|ccc@{}}
\toprule
 & \multicolumn{3}{c|}{\textbf{image-to-recipe}} & \multicolumn{3}{c}{\textbf{recipe-to-image}} \\  
Model & \small{R@1} & \small{R@5} & \small{R@10} & \small{R@1} & \small{R@5} & \small{R@10} \\ \midrule
\small{Baseline} & 40.0          & 67.0  &75.9       & 41.0       &67.3         &      75.9  \\ \midrule
\small{+ VSLP} &  41.1  & 67.5  & 76.1  & 42.4  & 68.1  & 76.5  \\ 
\small{+ VSLP \& Context} &  41.1  & 68.0    & 76.9 & 42.8 & 69.2  &  77.8 \\
\bottomrule
\end{tabular}
}
\caption{Ablation Study: Both VSLP and Context module bring significant improvement.}
\label{tab:VLPCook_ablation}
\end{table}

\subsection{Ablation Study of VLPCook}
\label{sec:ablation}

Here we present the ablation study for some design choices, on the 1k setup of Recipe1M test set: 

\paragraph{VLPCook (Sec.~\ref{sec:method}):} In Tab.~\ref{tab:VLPCook_ablation}, we show the effect of our contributions, mainly VLP and Context injection. We can notice that each one brings significant improvement compared to the baseline, as well as the combination of them.  

\paragraph{Local and Global Context (Sec.~\ref{sec:finetuning}):} In Tab.~\ref{tab:ablation_context}, we do an ablation on the type and the position of the injected context. We notice that using only the ingredients (Ing) or titles (Ttl) (lines 2 and 3 Tab.~\ref{tab:ablation_context})  outperforms the baseline (line 1) without any context. Moreover, using both contexts is always better, regardless of their position. We also show that the best configuration is by injecting the ingredients at the input to the visual encoder and the titles at the output (line 5). 

\begin{table}[h]
\small
\centering
\setlength{\tabcolsep}{5pt}
\resizebox{0.45\textwidth}{!}{
\begin{tabular}{@{}lccccccc@{}}
\toprule
& \multicolumn{2}{c}{Context} & \multicolumn{2}{c}{Position} & RSUM & RSUM & \multirow{2}{*}{RSUM}  \\ 
 & Ing & ttl & Input & Output & 1K & 10K &  \\ \midrule
 1 & \xmark &   \xmark     &  &        & 495.00 & 367.10 & 862.10   \\ \midrule
 2 & \cmark &        & \cmark &        & 500.54 & 371.43 & 871.97   \\ \midrule
 3 &     & \cmark &        & \cmark & 498.61 & 372.16 & 870.77 \\ \midrule
 4 & \cmark & \cmark & \cmark (ttl\&Ing) &        & 500.86   &  374.68  & 875.54   \\ 
 5 (ours) & \cmark & \cmark & \cmark (Ing) & \cmark (ttl) & 501.75  &  374.30 & \textbf{876.05}  \\ 
 6 & \cmark & \cmark & \cmark (ttl) & \cmark (Ing) & 501.79  &  372.44 & 874.23  \\ 
\bottomrule
\end{tabular}
}
\caption{\textbf{Ablation study on the context and injection position.} Local context (Ing) is better injected in the input of the ViT, and global one (ttl) in the output.}
\label{tab:ablation_context}
\end{table}

\paragraph{VSLP on the Recipe1M+ Dataset}
\label{sec:recipe1m+_exp}
Recipe1M+ is the largest dataset for food applications, however, to the best of our knowledge, there is no work, besides the work that introduced this dataset \cite{marin2019recipe1m+}, that consider it for cross-modal food retrieval. This might be due to, in addition to computation resources needed, the poor generalization from Recipe1M+ to Recipe1M as shown by the authors \cite{marin2019recipe1m+}. Here we try to leverage this dataset, and assess its benefit during pretraining. We pretrain several variants,  for 30 epochs on all the recipes of Recipe1M+ (after excluding those in the validation and test set of Recipe1M) following the same implementation details as Sec.~\ref{sec:method} (except training using only 2 GPUs), and then finetune these models on Recipe1M. The results of Tab.~\ref{tab:vlp_rec13m} show that Recipe1M+ is more effective than our IST, however, the latter contains only 1M images compared to 13M in the former, and the images and recipes are in the same distribution of those during finetuning. To fairly compare with IST, we also pretrain on Recipe1M+ by keeping only 10\% of the images (\emph{i.e} 1.3 images in average per recipe). Interestingly, we can notice from Tab.~\ref{tab:vlp_rec13m} that pretraining on IST leads to better results.

\begin{table}[h]
\small
\centering
\setlength{\tabcolsep}{5pt}
\resizebox{0.45\textwidth}{!}{
\begin{tabular}{@{}lccccccc@{}}
\toprule
\multirow{2}{*}{Model} & \multirow{2}{*}{VSLP} & \multicolumn{3}{c}{\textbf{image-to-recipe}} & \multicolumn{3}{c}{\textbf{recipe-to-image}} \\ 
 & & \small{R@1} & \small{R@5} & \small{R@10} & \small{R@1} & \small{R@5} & \small{R@10} \\ \midrule

\small{VLPCook w/o} & IST & 69.8 & 89.2 & 92.7 & 70.9 & 89.6 & 92.7  \\ 
\small{CLIP-ViT} & R1M+ & 71.0  & 89.3  & 92.7  & 71.9  & 89.6  & 92.7  \\ \midrule
 
 \multirow{2}{*}{\small{VLPCook}} & IST & 73.6  & 90.5    & 93.3  & 74.7   & 90.7  &  93.2  \\ 
 & R1M+ &  74.9   &  91.4    & 93.7   &  75.6   & 91.2  &  93.6  \\ \midrule
\small{VLPCook}  & R1M+ (1.3M Im.) &  73.4  & 90.7  & 93.2  & 73.8  & 90.8  & 93.1  \\

\bottomrule
\end{tabular}
}
\caption{\textbf{VSLP on our IST dataset vs on Recipe1M+ (R1M+).} Pretraining on R1M+ gives better results, however, for the same number of images, IST is a better choice.}
\label{tab:vlp_rec13m}
\end{table}

\subsection{Further Experiments}
\paragraph{Foundation Models in the Cooking Context.}
\label{sec:foundation_exp}
Best SoTA results on general benchmarks are currently obtained by finetuning foundation models, however, here we show that for tasks requiring more complex input, such as food retrieval, this paradigm lags significantly behind existing food models. To this end, we finetune on Recipe1M for cross-modal retrieval, considering  2 kinds of approaches; light fusion (CLIP) and heavy fusion (ALBEF) approaches.

\noindent\textbf{CLIP \cite{radford2021learning}:} Is trained contrastively on 400M of image-text pairs and consists of a ViT-Base/16 as image encoder and a transformer as text encoder.

\noindent\textbf{ALBEF \cite{li2021albef}:} Is trained using ITC, ITM and MLM losses on 14M images and their corresponding text. It consists of a ViT-Base/16 on the image side, a BERT on the text side, in addition to a multimodal decoder. 

For both models, we change the word embedding layer, the vocabulary, and maximum number of textual tokens to 300. We train for 120 epochs with the two losses; Adamine triplet with incremental margin, semantic regularization, and ITM (for ALBEF). We use Adam optimizer and learning rate of 1e-5 (for CLIP ViT we use lr of 1e-6) and a total batch size of 80 and 56 for CLIP and ALBEF respectively. 
Tab.~\ref{tab:Foundation} shows that CLIP and ALBEF give reasonable performance and outperform most of the baselines (Tab.~\ref{tab:main}). However, and contrary to other general benchmarks, their performance is still below SoTA food models.

\begin{table}[h]
\small
\centering
\setlength{\tabcolsep}{5pt}
\resizebox{0.4\textwidth}{!}{
\begin{tabular}{@{}lcccccc@{}}
\toprule
 & \multicolumn{3}{c}{\textbf{image-to-recipe}} & \multicolumn{3}{c}{\textbf{recipe-to-image}} \\ 
       Model              & \small{R@1}      & \small{R@5}    & \small{R@10}       & \small{R@1}     & \small{R@5}     & \small{R@10}    \\ \midrule
X-MRS \cite{guerrero2021cross_xmrs}     &     64.0    &    88.3   &    92.6    &     63.9   &    87.6    &      92.6    \\
H-T (ViT) \cite{salvador2021revamping}    &     64.2    &    89.1   &    93.4    &   64.5   &    89.3    &      93.8  \\
T-Food \cite{shukor2022transformertfood} & 68.2  & 87.9  &91.3  & 68.3  &87.8  & 91.5  \\\midrule
\small{CLIP} & 63.5 &85.4 & 90.0 & 64.1 & 85.8 & 90.1  \\ 
\small{ALBEF} & 61.0  & 84.7  & 89.9  &61.9  & 84.6  & 89.8  \\ 
\bottomrule
\end{tabular}
}
\caption{\textbf{Finetuning foundation models on Recipe1M.}}
\label{tab:Foundation}
\end{table}

\paragraph{Food Recognition.} Retrieval task is one of the best setups to evaluate cross-modal alignment, on the other hand, there is an established consensus in the community that cross-modal alignment significantly  helps solving multimodal downstream tasks. To echo this finding, we test the benefit of VLP for Food Recognition on Food101 \cite{food101} and the large ISIA Food500 \cite{min2020isia}. We compare SoTA food models to our VLPCooK pre-trained with VSLP, following the linear probe setup on top of frozen ViTs. Table .\ref{tab:classif} below shows very good results, e.g. we have a significant improvement in accuracy for Food Recognition. This shows the ability of our approach to generalize to other food tasks.

\setlength{\tabcolsep}{7pt}
\begin{table}[h]
\centering
\resizebox{0.48\textwidth}{!}{
\begin{tabular}{@{}l|cccc@{}}
\toprule
Food Recognition & ImageNet (ViT)  & H-T (ViT) &  VLPCook (ViT)    \\ \midrule
Food101 &  80.99     &  84.44   &   \textbf{89.14}        \\ 
ISIA Food500   &  52.34   &  57.562    &   \textbf{60.30}     \\

\bottomrule
\end{tabular}
}
\caption{Linear regression classification on the test sets of Food101 and ISIA Food500. Backbone (ViT) kept frozen.}
\label{tab:classif}
\end{table}

\subsection{Beyond Computational Cooking: Medical Domain}
Although stage 1 of our approach has been tailored for computational cooking tasks (stage 2), its design is more generally concerned with the processing of structured documents, and can be seamlessly adapted to other domains.  To support that, we consider structured data from very different domain, namely structured medical retrieval.

We experiment with Text-Image Retrieval for medical databases. We use the large scale ROCO dataset \cite{pelka2018radiologyroco} that consists of 81k radiology images and "reports" pairs, where the report contains a caption, keywords, Unified Medical Language Systems Concept Unique Identifiers (CUIs) and Semantic Types. We consider the list of keywords and Semantic Types as "ingredients", the caption as "instruction" and we extract the title from the caption (Sec.\ref{sec:structured_vlp}). 
Table \ref{tab:roco}, shows that our VSLP (VSLP) lead to additional $\sim$4 points of R@1 with respect to our baseline (VLPCook). This shows the broader impact of our approach and its benefits for domains and tasks requiring structured textual input.
\setlength{\tabcolsep}{7pt}
\begin{table}[h]
\centering
\resizebox{0.48\textwidth}{!}{
\begin{tabular}{@{}lc|cccccc@{}}
\toprule
    \multirow{2}{*}{Method} & \multirow{2}{*}{PT}              & \multicolumn{3}{c}{\textbf{image-to-text}} & \multicolumn{3}{c}{\textbf{text-to-image}} \\ 
               &  &   R@1      & R@5    & R@10    &    R@1     & R@5     & R@10    \\ \midrule

VLPCook &   $\varnothing$ &     14.53  &   38.20    &  51.71    &   15.08     &   39.03   &   51.83  \\
VLPCook &   VSLP &    \textbf{18.44}     &   \textbf{42.78}     &     \textbf{55.90}    & \textbf{17.95} &  \textbf{42.51} & \textbf{55.06}   \\
                 
\bottomrule
\end{tabular}
}
\caption{Comparison of different types of VLP on Image-Text Medical Retrieval on ROCO dataset.}
\label{tab:roco}
\end{table}

\section{Conclusion}
\label{sec:conclusion}
In this work, we show the benefits of  VSLP for Computational Cooking. We also, successfully leverage pretrained foundation models, to enrich the vision encoder with structured context. These contributions led to a new SoTA for Cross-Modal Food Retrieval. We show that this approach has a broader impact and can be adopted for other computational cooking applications or more general multimodal tasks, especially, those with complex input, such as Medical databases. An interesting follow up of this work, is to improve the textual structure extraction and going large scale in terms of pretraining data.

\section{Acknowledgments} The authors would like to thank Rémy Sun for fruitful discussion. This work was partly supported by ANR grant VISA DEEP (ANR-20-CHIA-0022), and
HPC resources of IDRIS under the allocation 2022-[AD011013415] and 2022-[A0121012449] made by GENCI.

\appendix

\section*{Appendix}
The Appendix is organized as follows; Sec.~\ref{sec:app_implem} elaborates on the implementation details, Sec.~\ref{sec:sota} presents the complete comparison of VLPCook with other SoTA approaches on Recipe1M and Recipe1M+ datasets. In section Sec.~\ref{sec:r1m+}, we conduct additional VSLP experiments. In Sec.~\ref{sec:missing_values}, we do a robustness analysis to missing recipe entities, where we show also the contribution of each of these entities for food retrieval. Finally, we show some qualitative examples on the extracted text (Sec.~\ref{sec:ste}) and the injected local and global context (Sec.~\ref{sec:lg_context}).

\section{Implementation details} 
\label{sec:app_implem}
\paragraph{VLP of VLPCook:} the model consists of a hierarchical transformer encoders and decoders on the recipe side, a ViT-B/16 \cite{vit} on the image side and a multimodal module \cite{shukor2022transformertfood}. For VLP, We start by pretraining this baseline with Adamine triplet (without semantic regularization losses) \cite{carvalho2018adamine} and ITM losses ($\lambda =1$), with learning rate (lr) 1e-5 and total batch size of 200, on 4 GPUs (50 per GPU) for 30 epochs. We pretrain on the 2M pairs of the IST dataset. Inspired by LiT \cite{zhai2022lit} we freeze the image encoder during this stage. 

\paragraph{Finetuning on Recipe1M:} in the second finetuning stage, we follow the implementation details of recent work \cite{shukor2022transformertfood}, mainly, batch size of 100, lr of 1e-5 (lr of 1e-6 for CLIP-ViT) and training for 120 epochs on the training set of Recipe1M. We optimize the model with the Adamine triplet (instance and semantic) with incremental margin (we start by a $\alpha_{inc}=0.05$ and increase it by $0.005$ each epoch until reaching $0.3$) and ITM objective ($\lambda =1$). The ViT is kept frozen for the first 20 epochs. Note that, we pretrain always with a ViT, even when we finetune with CLIP-ViT. We associate each image to 5 titles and 15 ingredients. These are extracted from the recipes of the training set of Recipe1M, using the CLIP-based retrieval approach. During training, we sample only 2 titles and 4 ingredients randomly in each batch. During Test we use all titles and ingredients. We concatenate the ingredients to the input of the ViT and the title to its output, before the linear projection to the latent space. The context is embedded by the first 2 layers of the BERT \cite{devlin2018bert} encoder, then linearly projected to obtain the context tokens, we find it beneficial to use separate BERT encoders for each context.

\paragraph{Finetuning on Recipe1M+:} for finetuning on Recipe1M+ \cite{marin2019recipe1m+}, we adopt the same implementation details as for Recipe1M, however, due to the large number of images (\emph{i.e.}, 13M) we extract the context from only 1 image for each recipe and use this context for all the other corresponding images. We finetune on 2 A100 GPUs, for 60 epochs, without the semantic triplet loss and keep the ViT frozen for the first 5 epochs.

\paragraph{Evaluation:} We follow other work and report recall@\{1, 5, 10\} (R@k) and their sum (RSUM), in addition to the median rank (medR) on the 1k and 10 setups, averaged over 10 and 5 runs respectively.

\paragraph{Image-Text Medical Retrieval.} We use the large scale ROCO dataset \cite{pelka2018radiologyroco} that consists of 81k radiology images and "reports" pairs, where the report contains a caption, keywords, Unified Medical Language Systems Concept Unique Identifiers (CUIs) and Semantic Types. We consider the list of keywords and Semantic Types as "ingredients", the caption as "instruction" and we extract the title from the caption as we did in our STE module. The results with standard VLP are reported from \cite{chen2022alignmed, m3ae}. We follow other approaches and evaluate on 2k pairs of the test set of ROCO. To ensure reproductibility, we average the results obtained on 4 different 2k subsets. Here we do not use the context module.

\setlength{\tabcolsep}{7pt}
\begin{table*}[t]
\centering
\resizebox{\textwidth}{!}{
\begin{tabular}{@{}l|cccc|cccc|cccc|cccc@{}}
\toprule
\multirow{3}{*}{} & \multicolumn{8}{c|}{\textbf{1k}}                                                    & \multicolumn{8}{c}{\textbf{10k}} \\ \cmidrule(l){2-17} 
                  & \multicolumn{4}{c|}{\textbf{image-to-recipe}} & \multicolumn{4}{c|}{\textbf{recipe-to-image}} & \multicolumn{4}{c|}{\textbf{image-to-recipe}} & \multicolumn{4}{c}{\textbf{recipe-to-image}} \\ \cmidrule(l){2-17} 
                  & medR     & R@1      & R@5    & R@10    & medR     & R@1     & R@5     & R@10    & medR     & R@1      & R@5    & R@10    & medR     & R@1     & R@5     & R@10    \\ \midrule
Salvador et al. \cite{Salvador_2017_CVPR_recipe1m}    & 5.2      & 24.0    & 51.0      &   65.0     &   5.1       & 25.0       &     52.0   &       65.0 & 41.9    & -     &  -     &   -     &      39.2    &      -  &     -   &    -    \\
Adamine \cite{carvalho2018adamine}   &    2.0      &   40.2     &   68.1     &   78.7  &    2.0      & 39.8    &  69.0     &    77.4     &      13.2    &   14.8     &    34.6    &    46.1 & 14.2     & 14.9    &   35.3    &  45.2   \\
R2GAN \cite{Zhu_2019_CVPR_r2gan}                  &     2.0     &     39.1    &   71.0    &  81.7      &  2.0        &    40.6    &    72.6    &       83.3 &  13.9        &   13.5      &   33.5    &     44.9   &     12.6     &   14.2     &   35.0     &   46.8     \\
MCEN \cite{fu2020mcen}               &   2.0       &   48.2      &    75.8   &      83.6  &    1.9      &   48.4     &  76.1      &    83.7    &  7.2        &      20.3   &    43.3   &   54.4     &     6.6     &   21.4     &  44.3      &    55.2    \\
ACME \cite{wang2019learning_acme}              &    1.0      &   51.8      &   80.2    &    87.5    &   1.0       &    52.8    &   80.2     &       87.6 &     6.7     &   22.9     &  46.8     &    57.9    &      6.0    &  24.4      &    47.9    &      59.0  \\
SN \cite{sentencebased}               &    1.0      & 52.7        &     81.7  &    88.9    &    1.0      &  54.1      &  81.8      &   88.9     &    7.0      &    22.1    &    45.9   &    56.9    &    7.0      &     23.4   &  47.3      &  57.9      \\
IMHF \cite{intra_inter}               &    1.0      & 53.2        &     80.7  &    87.6    &    1.0      &  54.1      &  82.4      &   88.2     &    6.2      &    23.4    &    48.2   &    58.4    &    5.8      &     24.9   &  48.3      &  59..4      \\
\textit{Wang et. al}  \cite{wang2021learning}               &    1.0      & 53.5        &     81.5  &    88.8    &    1.0      &  55.0      &  82.0      &   88.8     &    6.0      &    23.4    &    48.8   &    60.1    &    5.6      &     24.6   &  50.0      &  61.0      \\
SCAN \cite{wang2021cross_scan}               &    1.0      & 54.0        &     81.7  &    88.8    &    1.0      &  54.9      &  81.9      &   89.0     &    5.9      &    23.7    &    49.3   &    60.6    &    5.1      &     25.3   &  50.6      &  61.6      \\
HF-ICMA \cite{hybrid_fusion}               &    1.0      & 55.1        &     86.7  &    92.4    &    1.0      &  56.8      &  87.5      &   93.0     &    5.0      &    24.0    &    51.6   &    65.4    &    4.2      &     25.6   &  54.8      &  67.3      \\
MSJE \cite{xie2021learning}               &    1.0      & 56.5        &     84.7  &    90.9    &    1.0      &  56.2      &  84.9      &   91.1     &    5.0      &    25.6    &    52.1   &    63.8    &    5.0      &     26.2   &  52.5      &  64.1      \\

SEJE \cite{xie2021learning2}               &    1.0      & 58.1        &     85.8  &    92.2    &    1.0      &  58.5      &  86.2      &   92.3     &    4.2      &    26.9    &    54.0   &    65.6    &    4.0      &     27.2   &  54.4      &  66.1      \\

M-SIA \cite{multisubspace} &  1.0       &     59.3    &    86.3   &    92.6    &     1.0     &     59.8   &    86.7    &      92.8  &     4.0    &    29.2     &    55.0   &    66.2    &    4.0      &     30.3   &     55.6   &  66.5     \\

DaC \cite{dac}             &  1.0       &     60.2    &    84.0   &    89.7    &     -     &     -   &    -    &      -  &     4.0    &    30.0     &    56.5   &    67.0    &    -      &     -   &     -   &        \\

X-MRS \cite{guerrero2021cross_xmrs}             &  1.0       &     64.0    &    88.3   &    92.6    &     1.0     &     63.9   &    87.6    &      92.6  &     3.0    &    32.9     &    60.6   &    71.2    &    3.0      &     33.0   &     60.4   &  70.7       \\

H-T (ViT) \cite{salvador2021revamping}            &  1.0       &     64.2    &    89.1   &    93.4    &     1.0     &     64.5   &    89.3    &      \underline{93.8}  &     3.0    &    33.5     &    62.1   &    72.8    &    3.0      &     33.7   &     62.2   &  72.7       \\
\textit{Papadopoulos et al.} \cite{programfood}            &  1.0       &     66.9    &    \underline{90.9}   &   \textbf{95.1}    &     1.0     &     66.8   &    89.8    &   \textbf{94.6}  &    -    &    -     &    -   &    -    &    -      &     -   &     -   &  -       \\
T-Food (ViT) \cite{shukor2022transformertfood}          & 1.0 & 68.2  & 87.9  &91.3  & 1.0 & 68.3  &87.8  & 91.5 & 2.0      & 40.0          & 67.0  &75.9    &2.0   & 41.0       &67.3         &      75.9            \\ 
T-Food (CLIP-ViT)  \cite{shukor2022transformertfood}          &  1.0   &  72.3  & 90.7    & 93.4  & 1.0 & 72.6   & 90.6  &  93.4     &    2.0      & 43.4 & 70.7   & 79.7  & 2.0 &    44.6    & 71.2        & 79.7                 \\ \midrule
VLPCook           &  1.0    &  \underline{73.6}  & 90.5    & 93.3  & 1.0 & \underline{74.7}   & \underline{90.7}  &  93.2     &    2.0      & \underline{45.3}  & \underline{72.4}   & \underline{80.8}  & 2.0 &    \underline{46.4}    & \underline{73.1}        & \underline{80.9}                 \\ 
VLPCook  (R1M+)          & 1.0     &  \textbf{74.9}   &  \textbf{91.4}    & \underline{93.7}   & 1.0 &  \textbf{75.6}   & \textbf{91.2}  &  93.6     &   2.0     & \textbf{46.7}  & \textbf{73.3}   & \textbf{83.31}  & 2.0 &    \textbf{47.8}    & \textbf{74.1}   & \textbf{81.8}                 \\ 
\bottomrule
\end{tabular}
}
\caption{\textbf{Comparison with other work on the Recipe1M dataset.} medR ($\downarrow$), Recall@k ($\uparrow$) are reported on the Recipe1M test set. Our approaches (VLPCook) significantly outperform all existing work. Best metrics are in bold, and next best metrics are underlined.}
\label{tab:app_main_all}
\end{table*}

\setlength{\tabcolsep}{7pt}
\begin{table*}[t]
\centering
\resizebox{\textwidth}{!}{
\begin{tabular}{@{}l|cccc|cccc|cccc|cccc@{}}
\toprule
\multirow{3}{*}{} & \multicolumn{8}{c|}{\textbf{1k}}                                                    & \multicolumn{8}{c}{\textbf{10k}} \\ \cmidrule(l){2-17} 
                  & \multicolumn{4}{c|}{\textbf{image-to-recipe}} & \multicolumn{4}{c|}{\textbf{recipe-to-image}} & \multicolumn{4}{c|}{\textbf{image-to-recipe}} & \multicolumn{4}{c}{\textbf{recipe-to-image}} \\ \cmidrule(l){2-17} 
                  & medR     & R@1      & R@5    & R@10    & medR     & R@1     & R@5     & R@10    & medR     & R@1      & R@5    & R@10    & medR     & R@1     & R@5     & R@10    \\ \midrule
Marin et al. \cite{marin2019recipe1m+}    & 8.6      & 17.0     &  38.0     &   48.0     &   6.8 &   17.0    &    42.0  &     54.0 & -    & -     &  -     &   -     &      -    &      -  &     -   &    -    \\
VLPCook$^*$        &  2.0   &  45.2     &  75.9     &   84.0     &  2.0 &    47.3    &    77.6  &     85.3     &    9.2      &  18.0 & 40.7 & 52.2 & 8.0 &     19.8 & 43.4 & 55.0 \\

\bottomrule
\end{tabular}
}
\caption{\textbf{Comparison with other work on the Recipe1M+ dataset.} medR ($\downarrow$), Recall@k ($\uparrow$) are reported on the Recipe1M+ test set. Our approaches (VLPCook) significantly outperform all existing work. Best metrics are in bold, and next best metrics are underlined. All models are trained on the training set of Recipe1M+. $^*$ means without pretraining.}
\label{tab:app_main_recipe1m+_all}
\end{table*}

\section{Comparison with SoTA}
\label{sec:sota}
\begin{figure*}[h]
    \centering
    \includegraphics[width=\linewidth]{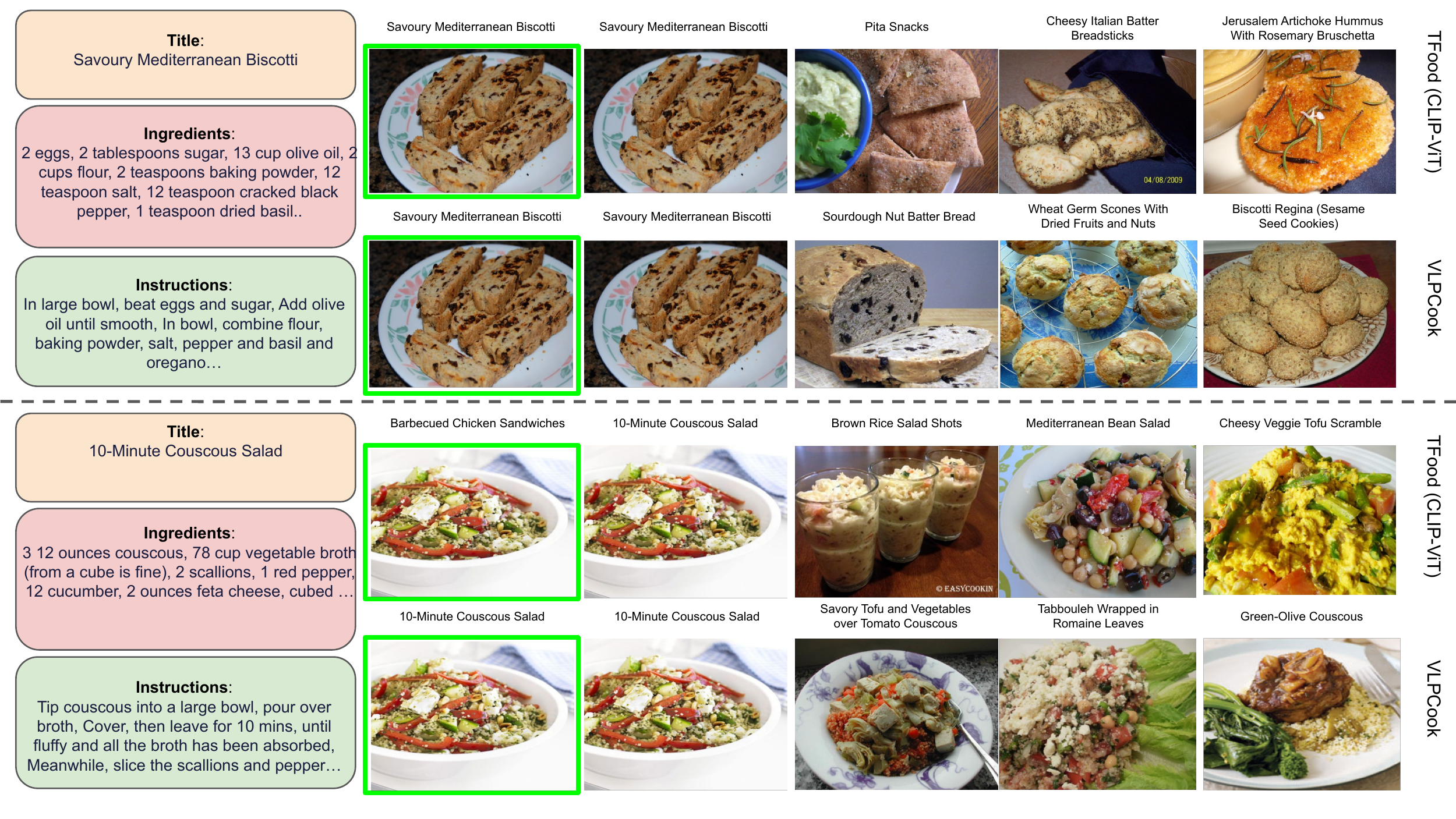}
     \caption{\textbf{Recipe to image qualitative results of VLPCook on the Recipe1M test set.} The image in green is the ground truth, followed by the top 4 retrieved images in order. For VLPCook, we can notice that all images semantically resemble the ground truth in addition to successfully retrieving the correct image.}
    \label{fig:app_qual_vs_tfood_1}
\end{figure*}

\begin{figure*}[t]
    \centering
    \includegraphics[width=\linewidth, height=9cm]{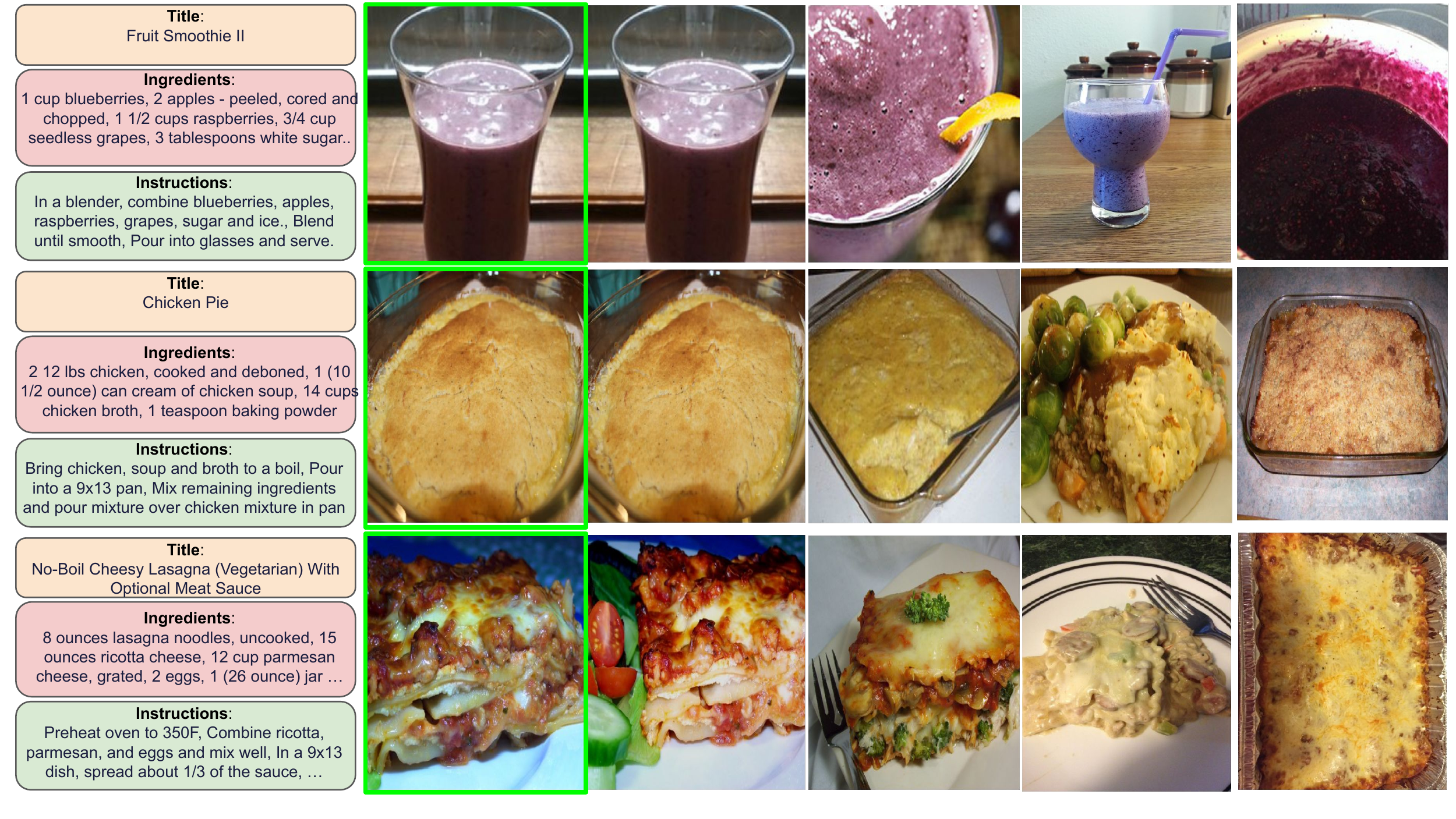}
     \caption{\textbf{Recipe to image qualitative results of VLPCook on the Recipe1M test set.} The image in green is the ground truth, followed by the top 4 retrieved images in order.}
    \label{fig:app_qual_1}
\end{figure*}

We compare VLPCook with other SoTA for Cross-Modal Food Retrieval. Tab.~\ref{tab:app_main_all} shows the results after finetuning on Recipe1M. We outperform other SoTA by a significant margin on the 1k (+2.1 R@1) and 10k (+1.9 R@1) setups. Pretraining on Recipe1M+ (R1M+) leads to additional improvements of +3 and +3.3 R@1 on the 1k and 10k setups respectively. We also show some qualitative results in  Fig.~\ref{fig:app_qual_vs_tfood_1} and ~\ref{fig:app_qual_1}.

The results of training on Recipe1M+ dataset are shown in Tab.~\ref{tab:app_main_recipe1m+_all}. We show the first interesting results on this challenging dataset, after the work \cite{marin2019recipe1m+} that introduced this dataset. Despite the large improvements, these results reveal the difficulty of this dataset, that could be interesting for devising more sophisticated approaches in the future.

\section{Additional VLP experiments}
\label{sec:vlp}

\paragraph{Vision and Structured-Language Pretraining Variants.} In Tab.~\ref{tab:vlp_ablation}, we compare different design choices for VSLP. The baseline is our implementation of TFood. We show the effectiveness of VSLP, especially the IST dataset, by the superiority of B+VSLP (ours) compared to B+VLP (w/o structure), which is a baseline that takes the same caption as title, ingredients and instructions, without extracting any structure. We also compare with pretraining all modules (B+VSLP (+Unfreeze Vis. Enc.)) and show that this degrades the performance. Finally, we use an object detector (VinVL \cite{zhang2021vinvl}) to extract the objects or local entities in the image, instead of our CLIP-based approach and show that both are competitive in the pretraining stage.

\begin{table}[h]
\small
\centering
\setlength{\tabcolsep}{5pt}
\resizebox{0.48\textwidth}{!}{
\begin{tabular}{@{}lcccccc@{}}
\toprule
 & \multicolumn{3}{c}{\textbf{image-to-recipe}} & \multicolumn{3}{c}{\textbf{recipe-to-image}} \\  
Model & \small{R@1} & \small{R@5} & \small{R@10} & \small{R@1} & \small{R@5} & \small{R@10} \\ \midrule
\small{Baseline (B)} & 68.2 & 87.9 & 91.3 & 68.3 & 87.8 & 91.5  \\ \midrule
\small{B + VLP (w/o strcuture)} & 67.2  & 87.3  & 91.0  & 67.5  & 87.5  & 91.1  \\ 
\small{B + VSLP (Unfreeze Vis. Enc.)} & 67.6  & 87.3  & 91.3  & 67.6  & 87.2  & 90.9  \\
\small{B + VSLP (w/ VinVL tags)} & 68.8  & 88.3  & 91.8  & 69.9  & 88.3  & 91.7  \\ \midrule
\small{B + VSLP (ours)} & 69.5  & 88.0  & 91.4  &69.7  & 88.1  & 91.5  \\ 
\bottomrule
\end{tabular}
}
\caption{\textbf{Ablation study on VSLP.} Different variants of VSLP.}
\label{tab:vlp_ablation}
\end{table}

\paragraph{VLP of Existing Food Models.}
\label{sec:vlp_exp}
We now validate that VLP consistently improves a wide variety of existing food models. We experiments with 2 kinds of approaches; with standard transformer (\emph{e.g}, BERT)  such X-MRS \cite{guerrero2021cross_xmrs} and VLPCook-B (BERT) (our Baseline where we replace the recipe encoder by a BERT) and with hierarchical transformers such as TFood.  We do not change the training procedure for these methods, the only difference is in the pretraining stage, or initialization. 
We train on the 2M pairs. The BERT-based models are trained with image captions (training on IST can be found in the appendix) and those with hierarchical transformers with our transformed datasets (structured text). Results are reported in Tab.~\ref{tab:vlp}, that shows a consistent improvement for all SoTA with VLP. This validates the benefit of using VLP for cross-modal food retrieval and shows the effectiveness of our approach to transform captions to structured text.

\begin{table}[h]
\small
\centering
\setlength{\tabcolsep}{5pt}
\resizebox{0.45\textwidth}{!}{
\begin{tabular}{@{}lccccccc@{}}
\toprule
\multirow{2}{*}{Model} & \multirow{2}{*}{VLP} & \multicolumn{3}{c}{\textbf{image-to-recipe}} & \multicolumn{3}{c}{\textbf{recipe-to-image}} \\ 
 & & \small{R@1} & \small{R@5} & \small{R@10} & \small{R@1} & \small{R@5} & \small{R@10} \\ \midrule
 \multirow{2}{*}{\small{XMRS}} & \xmark & 60.9  & 85.6  & 90.8  & 61.2  & 85.9  & 91.0  \\ 
& \cmark & 61.8  & 86.3  & 91.6  &62.7  & 86.7  & 91.7  \\ \midrule
\small{VLPCook-B} & \xmark & 61.4  & 84.1  & 88.8  & 61.3  & 84.3  & 89.0  \\ 
 \small{(BERT)} & \cmark & 63.4  & 85.3  & 89.5  &63.2  & 85.3  & 89.7  \\ \midrule
\multirow{2}{*}{\small{TFood}} & \xmark & 68.2 & 87.9 & 91.3 & 68.3 & 87.8 & 91.5  \\ 
 & \cmark & 69.5  & 88.0  & 91.4  &69.7  & 88.1  & 91.5  \\ 

\bottomrule
\end{tabular}
}
\caption{\textbf{Results of VLP with existing food approaches.} We see consistent improvement with VLP.}
\label{tab:vlp}
\end{table}

\paragraph{Pretraining on Recipe1M+}
\label{sec:r1m+}
In this section, we analyse the influence of the number of images and recipes for VSLP. We pretrain on different subset sizes of Recipe1M+ dataset. From Tab.~\ref{tab:vlp_rec13m_sizes}, we can notice that there is a significant improvement when adding more images. Interestingly, for comparable number of images, pretraining on IST gives better performance. 

When reducing the number of recipes in Tab.~\ref{tab:vlp_rec13m_sizes_recipe}, we can notice also a significant degradation. 

Interestingly, reducing the number of recipes or images to half, leads to comparable results (73.9 R@1 for 6.5M images in Tab.~\ref{tab:vlp_rec13m_sizes} or 0.45M recipes in Tab.~\ref{tab:vlp_rec13m_sizes_recipe}).

\begin{table}[h]
\small
\centering
\setlength{\tabcolsep}{5pt}
\resizebox{0.45\textwidth}{!}{
\begin{tabular}{@{}cccccccc@{}}
\toprule
Pretraining & \multirow{2}{*}{\# Images} & \multicolumn{3}{c}{\textbf{image-to-recipe}} & \multicolumn{3}{c}{\textbf{recipe-to-image}} \\ 
Dataset & & \small{R@1} & \small{R@5} & \small{R@10} & \small{R@1} & \small{R@5} & \small{R@10} \\ \midrule

\small{IST} & 1M & 73.6  & 90.5    & 93.3  & 74.7   & 90.7  &  93.2  \\ \midrule
\small{R1M+} & 1.3M & 73.4  & 90.7  & 93.2  & 73.8  & 90.8  & 93.1  \\ 
\small{R1M+} & 6.5M & 73.9  & 91.0  & 93.6  & 74.8  & 91.2  & 93.7  \\
\small{R1M+} & 13M & 74.9   &  91.4    & 93.7   &  75.6   & 91.2  &  93.6  \\
\bottomrule
\end{tabular}
}
\caption{VLPCook pretrained on IST and subsets of Recipe1M+ with different number of images; 1.3M (10\%), 6.5M (50\%) and 13M (100\%).}
\label{tab:vlp_rec13m_sizes}
\end{table}

\begin{table}[h]
\small
\centering
\setlength{\tabcolsep}{5pt}
\resizebox{0.45\textwidth}{!}{
\begin{tabular}{@{}ccccccccc@{}}
\toprule
Pretraining & \multirow{2}{*}{\% of Images} & \multirow{2}{*}{\# Recipes} & \multicolumn{3}{c}{\textbf{image-to-recipe}} & \multicolumn{3}{c}{\textbf{recipe-to-image}} \\ 
Dataset & & & \small{R@1} & \small{R@5} & \small{R@10} & \small{R@1} & \small{R@5} & \small{R@10} \\ \midrule
\small{R1M+} & 10\% & 0.9M & 73.4  & 90.7  & 93.2  & 73.8  & 90.8  & 93.1  \\ 
\small{R1M+} & 10\% & 0.45M & 72.7  & 90.4  & 93.5  & 73.5  & 90.8  & 93.6  \\ \midrule
\small{R1M+} & 100\% & 0.9M & 74.9   &  91.4    & 93.7   &  75.6   & 91.2  &  93.6  \\
\small{R1M+} & 100\% & 0.45M & 73.9   &  90.8    & 93.4   &  74.6   & 91.0  &  93.5  \\
\bottomrule
\end{tabular}
}
\caption{VLPCook pretrained on IST and subsets of Recipe1M+ with different number of recipes; 0.45M (50\%), and 0.9M (100\%).}
\label{tab:vlp_rec13m_sizes_recipe}
\end{table}

\section{Robustness to missing recipe entities}
\label{sec:missing_values}

Here we analyse how much our model is robust against missing recipe entities. In addition, this will help to understand the importance of each element, and how much they contribute to find the right visual representation. This may also have some important applications in several scenarios (\emph{e.g.} in case we have a specific ingredients, and we are wondering what can we make from them). The results are shown in Tab.~\ref{tab:missing_values}. We can notice that the most important elements are the ingredients, then the instructions and finally the title. Compared to TFood (CLIP-ViT) \cite{shukor2022transformertfood}, in general we are more robust, except for missing ingredients. This indicates that our model rely heavily on the ingredients to find the image which might be caused by the local context (ingredients injected in the vision encoder) that might steer the model to focus more on the ingredients.

\begin{table}[h]
\small
\centering
\setlength{\tabcolsep}{5pt}
\resizebox{0.45\textwidth}{!}{
\begin{tabular}{@{}lccccccc@{}}
\toprule
Missing & \multirow{2}{*}{Model} & \multicolumn{3}{c}{\textbf{image-to-recipe}} & \multicolumn{3}{c}{\textbf{recipe-to-image}} \\ 
 entity & & \small{R@1} & \small{R@5} & \small{R@10} & \small{R@1} & \small{R@5} & \small{R@10} \\ \midrule
 \multirow{2}{*}{\small{Ttl}} & \small{TFood (CLIP-ViT)} & 65.6  & 87.8  & 91.8  & 64.2  & 86.9  & 91.1  \\ 
& \small{VLPCook} & 68.6  & 88.1  & 92.0  & 68.4  & 87.6  & 91.3  \\ \midrule
\multirow{2}{*}{\small{Ing}} & \small{TFood (CLIP-ViT)} & 40.6  & 69.6  & 78.6  & 30.4  & 57.5  & 67.3  \\ 
 & \small{VLPCook} & 36.5  & 65.7  & 75.3  &24.9  & 52.4  & 63.9  \\ \midrule
\multirow{2}{*}{\small{Ins}} & \small{TFood (CLIP-ViT)} & 62.1 & 84.9 & 90.1 & 57.5 & 82.3 & 88.2  \\ 
 & \small{VLPCook} & 64.1  & 85.7  & 90.0  & 62.0  & 83.8  & 88.6  \\ \midrule
\multirow{2}{*}{\small{Ttl+Ins}} & \small{TFood (CLIP-ViT)} & 45.5 & 72.3 & 80.6 & 34.5 & 61.8 & 72.3  \\ 
 & \small{VLPCook} & 51.0  & 76.9  & 83.6  & 42.9  & 69.6  & 78.1  \\

\bottomrule
\end{tabular}
}
\caption{\textbf{Robustness to missing recipe entities.} The ingredients contribute more to finding the corresponding example, then the instructions, and finally the title.}
\label{tab:missing_values}
\end{table}

\section{Structured Text Extraction (STE)}
\label{sec:ste}
\begin{figure*}[h]
    \centering
    \includegraphics[width=\linewidth]{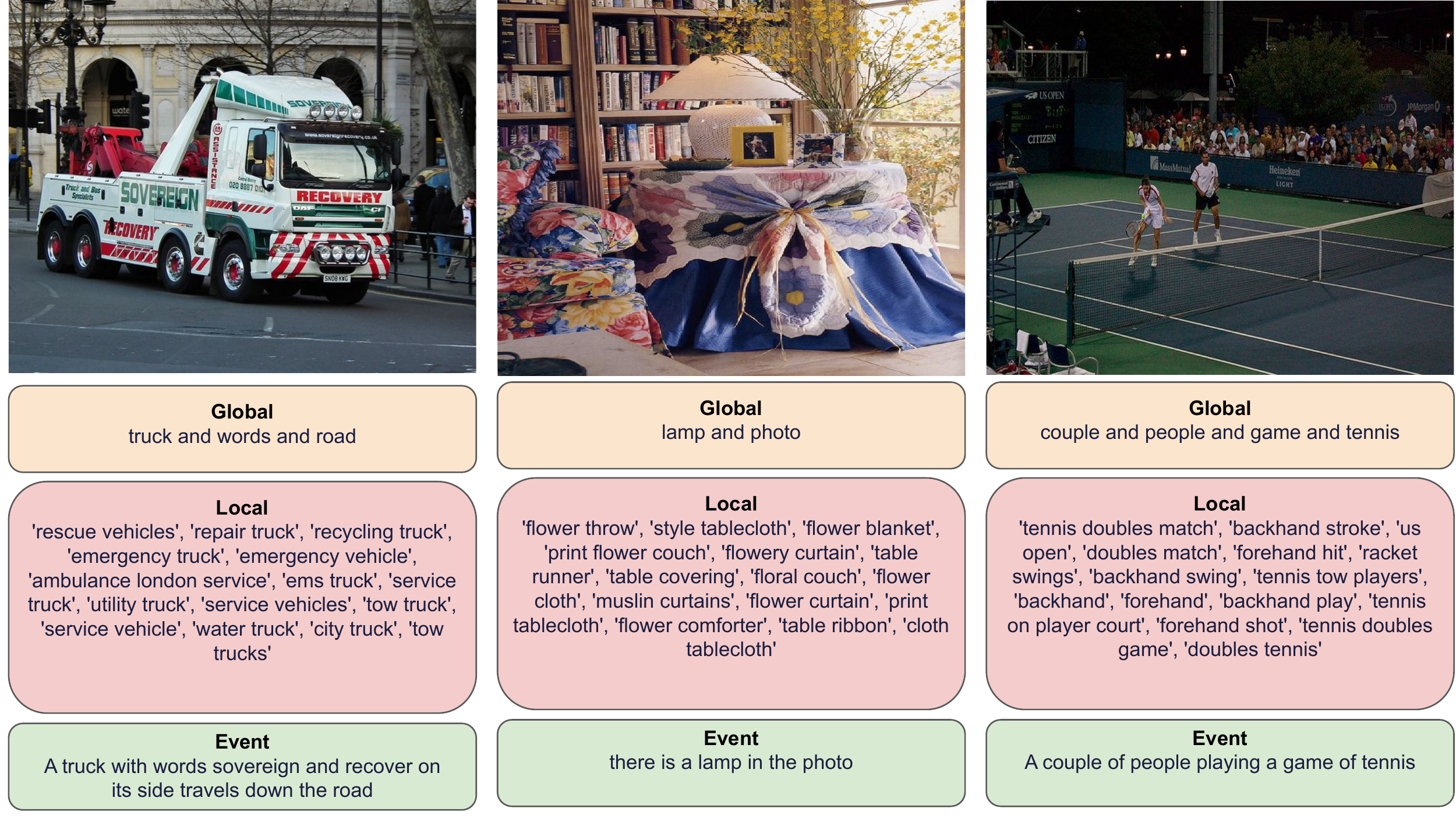}
     \caption{\textbf{Illustration of the structured text, extracted by the STE module.} For each image, we extract a global information using SGP , local information using CLIP-based retrieval and the event which is simply the caption.}
    \label{fig:ste_vis}
\end{figure*}

We illustrate in Fig.~\ref{fig:ste_vis} some qualitative examples of the structured text, obtained after transforming image captions using the STE module. We can see that the local elements are related mostly to the center of the image, describe the main or central object, and redundant. While such extracted information proved to be useful for food retrieval, devising other approaches that extracts information about all seen objects, with richer details, can help for tasks requiring more complex reasoning.  

\section{Local and Global Textual Concepts}
\label{sec:lg_context}
\begin{figure*}[h]
    \centering
    \includegraphics[width=\linewidth]{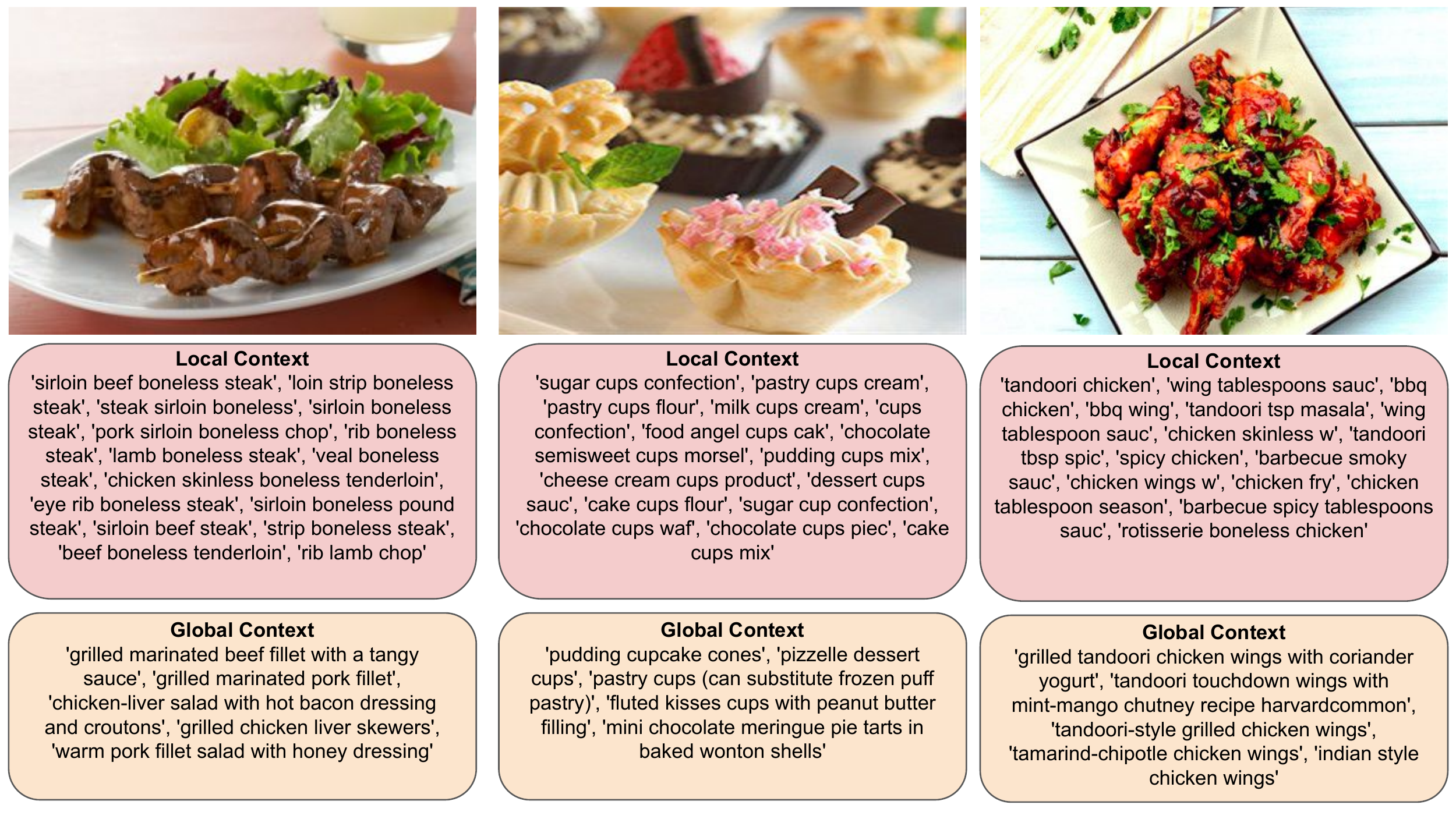}
     \caption{\textbf{Illustration of the local and global concepts}. Both concepts are extracted using CLIP-based retrieval. The local concepts consists of ingredients, and the global ones as recipe titles.}
    \label{fig:local_global_vis}
\end{figure*}

Fig.~\ref{fig:local_global_vis} shows the extracted context associated with each image. We successfully extract relevant contexts describing the recipe. However, we have also the redundancy in the local context, which might be due to the biases in the CLIP to the central objects in the image.

\clearpage

{\small
\bibliographystyle{ieee_fullname}
\bibliography{VLP_cooking_arxiv}
}

\end{document}